%% file: main.tex
\documentclass[10pt,twocolumn,letterpaper]{article}

\usepackage{cvpr}              

\usepackage{graphicx}
\usepackage{amsmath}
\usepackage{amssymb}
\usepackage{booktabs}
\usepackage[table]{xcolor}

\usepackage[pagebackref,breaklinks,colorlinks]{hyperref}

\usepackage[capitalize]{cleveref}
\crefname{section}{Sec.}{Secs.}
\Crefname{section}{Section}{Sections}
\Crefname{table}{Table}{Tables}
\crefname{table}{Tab.}{Tabs.}

\usepackage{subfiles} 

\def\cvprPaperID{6626} 
\def\confName{CVPR}
\def\confYear{2023}

\begin{document}

\title{StyleRF: Zero-shot 3D Style Transfer of Neural Radiance Fields\vspace{-0.8em}}

\author{
Kunhao Liu$^{1}$
\quad
Fangneng Zhan$^{2}$
\quad
Yiwen Chen$^{1}$ 
\quad
Jiahui Zhang$^1$ 
\\[0.5mm]
Yingchen Yu$^1$
\quad
Abdulmotaleb El Saddik$^{3,5}$
\quad
Shijian Lu$^{1*}$
\quad
Eric Xing$^{4,5}$
\\[2mm]
{\small $^1$Nanyang Technological University\quad$^2$Max Planck Institute for Informatics}\\[0.1mm]
{\small $^3$University of Ottawa\quad$^4$Carnegie Mellon University\quad$^5$MBZUAI}
\vspace{-1em}
}



\twocolumn[{
    \renewcommand\twocolumn[1][]{#1}
    \maketitle
    \begin{center}
        \centering
        \captionsetup{type=figure}
        \includegraphics[width=1.\textwidth]{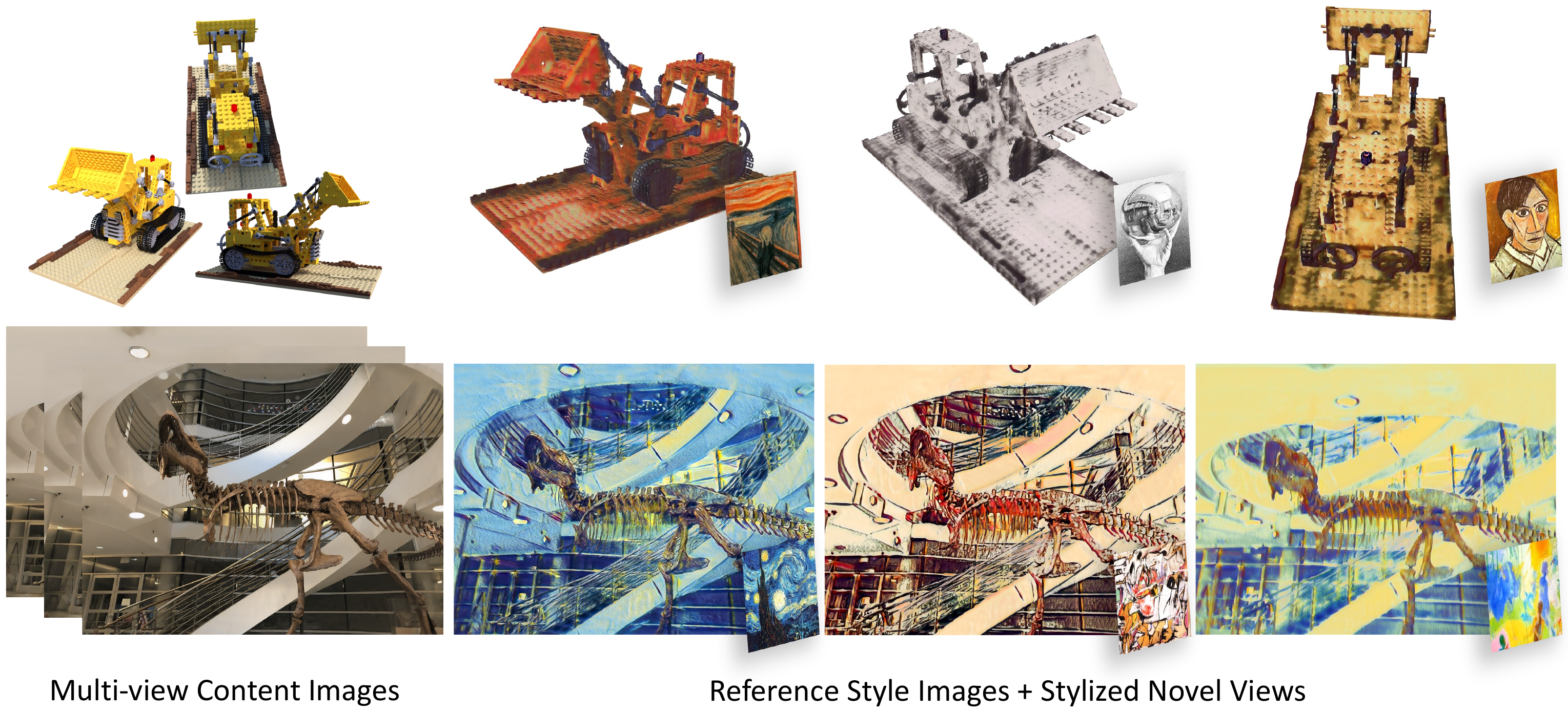}
        \captionof{figure}{\textbf{Zero-shot 3D Style Transfer.} Given a set of \textit{Multi-view Content Images} of a 3D scene, StyleRF can transfer arbitrary \textit{Reference Styles} to the 3D scene in a zero-shot manner, rendering high-quality \textit{Stylized Novel Views} with superb multi-view consistency.
        }
        \label{fig:teaser}
    \end{center}
}]

\def\thefootnote{*}\footnotetext{Shijian Lu is the corresponding author.}



\subfile{sections/abstract}


\subfile{sections/introduction}

\subfile{sections/relatedwork}

\section{Method}

\subfile{sections/featuregrid}

\subfile{sections/styletransfer}

\subfile{sections/training}

\subfile{sections/results}

\subfile{sections/studies}

\subfile{sections/conclusion}

{\small
\bibliographystyle{ieee_fullname}
\bibliography{egbib}
}

\end{document}

%% file: sections/abstract.tex
\begin{abstract}
3D style transfer aims to render stylized novel views of a 3D scene with multi-view consistency. However, most existing work suffers from a three-way dilemma over accurate geometry reconstruction, high-quality stylization, and being generalizable to arbitrary new styles. We propose StyleRF (Style Radiance Fields), an innovative 3D style transfer technique that resolves the three-way dilemma by performing style transformation within the feature space of a radiance field. StyleRF employs an explicit grid of high-level features to represent 3D scenes, with which high-fidelity geometry can be reliably restored via volume rendering. In addition, it transforms the grid features according to the reference style which directly leads to high-quality zero-shot style transfer. StyleRF consists of two innovative designs. The first is sampling-invariant content transformation that makes the transformation invariant to the holistic statistics of the sampled 3D points and accordingly ensures multi-view consistency. The second is deferred style transformation of 2D feature maps which is equivalent to the transformation of 3D points but greatly reduces memory footprint without degrading multi-view consistency. Extensive experiments show that StyleRF achieves superior 3D stylization quality with precise geometry reconstruction and it can generalize to various new styles in a zero-shot manner. Project website: \url{https://kunhao-liu.github.io/StyleRF/}
\end{abstract}
\vspace{-1em}

%% file: sections/introduction.tex
\section{Introduction}

Given a set of multi-view images of a 3D scene and an image capturing a target style, 3D style transfer aims to generate novel views of the 3D scene that have the target style consistently across the generated views (\cref{fig:teaser}). Neural style transfer has been investigated extensively, and state-of-the-art methods allow transferring arbitrary styles in a zero-shot manner. However, most existing work focuses on style transfer across 2D images~\cite{gatys2016image, johnson2016perceptual, huang2017arbitrary} but cannot extend to a 3D scene that has arbitrary new views. Prior studies~\cite{huang2022stylizednerf, nguyen2022snerf, huang2021learning,mu20223d} have shown that naively combining 3D novel view synthesis and 2D style transfer often leads to multi-view inconsistency or poor stylization quality, and 3D style transfer should optimize novel view synthesis and style transfer jointly.

However, the current 3D style transfer is facing a three-way dilemma over accurate geometry reconstruction, high-quality stylization, and being generalizable to new styles. Different approaches have been investigated to resolve the three-way dilemma. For example, multiple style transfer~\cite{huang2022stylizednerf, fan2022unified} requires a set of pre-defined styles but cannot generalize to unseen new styles. Point-cloud-based style transfer~\cite{huang2021learning,mu20223d} requires a pre-trained depth estimation module that is prone to inaccurate geometry reconstruction. Zero-shot style transfer with neural radiance fields (NeRF)~\cite{chiang2022stylizing} cannot capture detailed style patterns and textures as it implicitly injects the style information into neural network parameters. Optimization-based style transfer~\cite{zhang2022arf, hollein2022stylemesh, nguyen2022snerf} suffers from slow optimization and cannot scale with new styles.

In this work, we introduce \textbf{StyleRF} to resolve the three-way dilemma by performing style transformation in the feature space of a radiance field. A radiance field is a continuous volume that can restore more precise geometry than point clouds or meshes. In addition, transforming a radiance field in the feature space is more expressive with better stylization quality than implicit methods \cite{chiang2022stylizing}, and it can also generalize to arbitrary styles. We construct a 3D scene representation with a grid of deep features to enable feature transformation. In addition, multi-view consistent style transformation in the feature space could be achieved by either transforming the whole feature grid or transforming the sampled 3D points. We adopt the latter as the former incurs much more computational cost during training to stylize the whole feature grid in every iteration, whereas the latter can reduce computational cost through decreasing the size of training patch and the number of sampled points.  However, applying off-the-shelf style transformations to a batch of sampled 3D points impairs the multi-view consistency as they are conditioned on the holistic statistics of the batch. Beyond that, transforming every sampled 3D point is memory-intensive since NeRF needs to query hundreds of sampled points along each ray for rendering a single pixel.

We decompose the style transformation into sampling-invariant content transformation (SICT) and deferred style transformation (DST), the former eliminating the dependency on holistic statistics of sampled point batch and the latter deferring style transformation to 2D feature maps for better efficiency. In SICT, we introduce volume-adaptive normalization that learns the mean and variance of the whole volume instead of computing them from a sampled batch. In addition, we apply channel-wise self-attention to transform each 3D point independently to make it conditioned on the feature of that point regardless of the holistic statistics of the sampled batch. In DST, we defer the style transformation to the volume-rendered 2D feature maps based on the observation that the style transformation of each point is the same. By formulating the style transformation by pure matrix multiplication and adaptive bias addition, transforming 2D feature maps is mathematically equivalent to transforming 3D point features but it saves computation and memory greatly. Thanks to the memory-efficient representation of 3D scenes and deferred style transformation, our network can train with \(256 \times 256\) patches directly without requiring sub-sampling like previous NeRF-based 3D style transfer methods\cite{huang2022stylizednerf, chiang2022stylizing, fan2022unified}.

The contributions of this work can be summarized in three aspects. \textit{First}, we introduce StyleRF, an innovative zero-shot 3D style transfer framework that can generate zero-shot high-quality 3D stylization via style transformation within the feature space of a radiance field. \textit{Second}, we design sampling-invariant content transformation and deferred style transformation, the former achieving multi-view consistent transformation by eliminating dependency on holistic statistics of sampled point batch while the latter greatly improves stylization efficiency by deferring style transformation to 2D feature maps. \textit{Third}, extensive experiments show that StyleRF achieves superior 3D style transfer with accurate geometry reconstruction, high-quality stylization, and great generalization to new styles.

\begin{figure*}
    \centering
    \includegraphics[scale=.8]{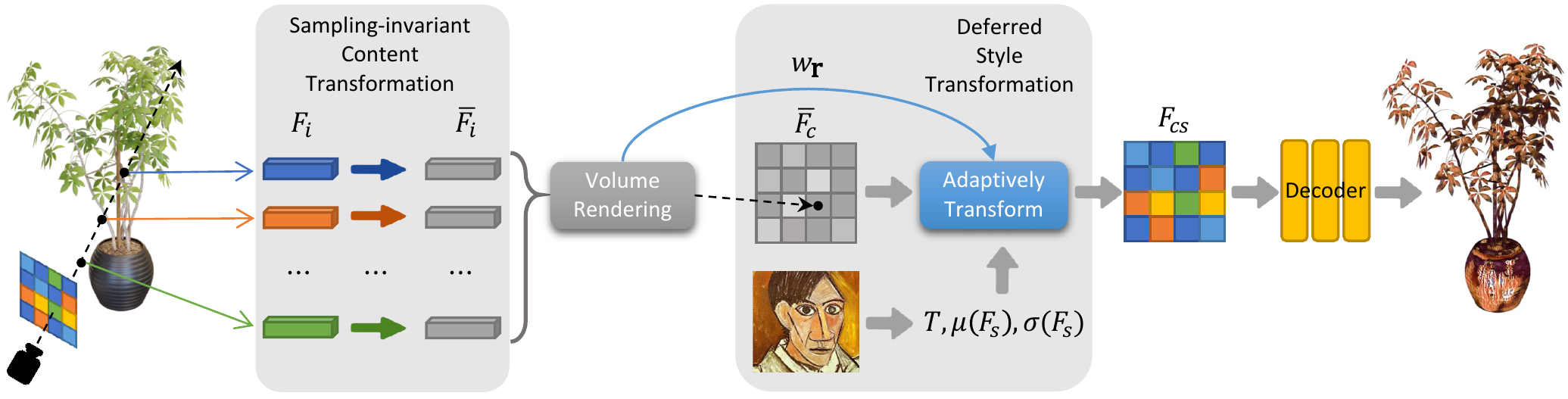}
    \caption{\textbf{The framework of StyleRF.}  For a batch of sampled points along a ray \( \textbf{r} \), the corresponding features \(F_i, i \in [1,2,...,N]\) are first extracted, each of which is transformed to \( \bar{F_i} \) independently via \textit{Sampling-Invariant Content Transformation}, regardless of the holistic statistics of the point batch. \( \bar{F_i} \) is then transformed to a feature map \( \bar{F_c} \) via \textit{Volume Rendering}. After that, the \textit{Deferred Style Transformation} transforms \( \bar{F_c} \) to the feature map \(F_{cs}\) adaptively using the sum weight of the sampled points \( w_{\textbf{r}} \) along the ray \( \textbf{r} \) and the style information \(T,\mu(F_s)\), and \(\sigma(F_s)\). Finally, a stylized novel view is generated via a CNN decoder.}
    \vspace{-1em}
    \label{fig:overview}
\end{figure*}

%% file: sections/relatedwork.tex
\section{Related Work}

\noindent \textbf{Neural scene representations.} 
3D scene representation has been extensively studied in recent years with different ways of representations such as volumes~\cite{ji2017surfacenet,wu20153d,seitz1999photorealistic, qi2016volumetric,kutulakos2000theory}, point clouds~\cite{achlioptas2018learning,qi2017pointnet}, meshes\cite{kanazawa2018learning, wang2018pixel2mesh}, depth maps~\cite{huang2018deepmvs, liu2015learning}, and implicit functions~\cite{chen2019learning,mescheder2019occupancy, niemeyer2020differentiable,yariv2020multiview}. These methods adopt differentiable rendering which enables model optimization by using 2D multi-view images. Among them, Neural Radiance Field (NeRF)~\cite{mildenhall2021nerf} can render a complex 3D scene with high fidelity and accurate geometry. It represents scenes with an implicit coordinate function that maps each 3D coordinate to a density value and a color value, and employs volume rendering to generate images of novel views. However, the implicit coordinate function is represented by a large multilayer perceptron (MLP) that is often hard to optimize and slow to infer. Serval studies adopt a hybrid representation \cite{chan2022efficient,fridovich2022plenoxels,sun2022direct,muller2022instant, chen2022tensorf,peng2020convolutional,martel2021acorn,liu2020neural,devries2021unconstrained,zhan2023general} to speed up the reconstruction and rendering. They employ explicit data structures such as discrete voxel grids \cite{sun2022direct,fridovich2022plenoxels}, decomposed tensors\cite{chen2022tensorf, chan2022efficient,fridovich2023k}, hash maps \cite{muller2022instant}, etc. to store features or spherical harmonics, enabling fast convergence and inference.

Although most existing work extracts features as middle-level representations of scenes, the extracted features are usually an intermediate output of neural networks which have little semantic meanings and are not suitable for the style transfer task. We introduce decomposed tensors~\cite{chen2022tensorf} to store high-level features extracted by pre-trained CNNs, which enables transformations in feature space as well as efficient training and inference. Though~\cite{niemeyer2021giraffe,chan2022efficient,gu2021stylenerf} also render feature maps instead of RGB maps, they are computationally intensive and usually work with low-resolution feature maps. StyleRF can instead render full-resolution feature maps (the same as the output RGB images) efficiently and it uses high-level features largely for transformation only.

\noindent \textbf{Neural style transfer.} 
Neural style transfer aims at rendering a new image that contains the content structure of one image and the style patterns of another. The seminal work in~\cite{gatys2016image} shows that multi-level feature statistics extracted from intermediate layers of pre-trained CNNs could be used as a representation of the style of an artistic image, but it treats style transfer as a slow and iterative optimization task. \cite{johnson2016perceptual,huang2017arbitrary,li2017universal, wu2021styleformer,sheng2018avatar,liu2021adaattn,park2019arbitrary,deng2020arbitrary,li2019learning} utilize feed-forward networks to approximate the optimization procedure to speed up rendering. Among them, \cite{huang2017arbitrary,li2017universal, wu2021styleformer,sheng2018avatar,liu2021adaattn,park2019arbitrary,deng2020arbitrary,li2019learning} can achieve zero-shot style transfer by applying transformations to the high-level features extracted by pre-trained CNNs, where the feature transformations can be achieved by matching second-order statistics \cite{huang2017arbitrary,li2017universal}, linear transformation \cite{li2019learning,wu2021styleformer} , self-attention transformation \cite{liu2021adaattn, park2019arbitrary,deng2020arbitrary}, etc. 
Video style transfer extends style transfer to videos for injecting target styles consistently across adjacent video frames. Several studies leverage optical flow~\cite{chen2017coherent, huang2017real,ruder2018artistic, wang2020consistent,ReReVST2020} as temporal constraints to estimate the movement of video contents. They can produce smooth videos, but have little knowledge of the underlying 3D geometry and cannot render consistent frames in arbitrary views \cite{mu20223d,huang2021learning}.

Huang et al. first tackle 
stylizing complex 3D scenes~\cite{huang2021learning}. They construct a 3D scene by back-projecting image features into the 3D space to form a point cloud and then perform style transformation on the features of 3D points. Their method can achieve zero-shot style transfer, but requires 
an error-prone pre-trained depth estimator to model scene geometry. \cite{mu20223d} also constructs a point cloud for stylization but it mainly focuses on monocular images. Instead, \cite{huang2022stylizednerf,chiang2022stylizing,zhang2022arf,nguyen2022snerf, fan2022unified,chen2022upstnerf} use NeRF\cite{mildenhall2021nerf} as the 3D representation which can reconstruct scene geometry more faithfully. 
\cite{chen2022upstnerf} is a photorealistic style transfer method that can only transfer the color tone of style images.
\cite{zhang2022arf,nguyen2022snerf} achieve 3D style transfer via optimization and can produce visually high-quality stylization, but they 
require a time-consuming optimization procedure for every reference style. \cite{huang2022stylizednerf, fan2022unified} employ latent codes to represent a set of pre-defined styles, but cannot generalize to unseen styles. \cite{chiang2022stylizing} can achieve arbitrary style transfer by implicitly instilling the style information into MLP parameters. 
However, it can only transfer the color tone of style images but cannot capture detailed style patterns. StyleRF can transfer arbitrary style in a zero-shot manner, and it can capture style details such as strokes and textures as well.

%% file: sections/featuregrid.tex
The overview of StyleRF is shown in \cref{fig:overview}. For a batch of sampled points along a ray \( \textbf{r} \), the corresponding features \(F_i, i \in [1,2,...,N]\) are first extracted from the feature grid described in \cref{sec:grid}, each of which is transformed to \( \bar{F_i} \) independently via \textit{Sampling-Invariant Content Transformation (SICT)} described in \cref{sec:SICT}, regardless of the holistic statistics of the point batch. \( \bar{F_i} \) is then transformed to a feature map \( \bar{F_c} \) via \textit{Volume Rendering}. After that, the \textit{Deferred Style Transformation (DST)} described in \cref{sec:DST} transforms \( \bar{F_c} \) to the feature map \(F_{cs}\) adaptively using the sum weight of the sampled points \( w_{\textbf{r}} \) along the ray \( \textbf{r} \) and the style information \(T,\mu(F_s),\) and \(\sigma(F_s)\). Finally, a stylized novel view is generated via a CNN decoder.

\subsection{Feature Grid 3D Representation}
\label{sec:grid}

To model a 3D scene with deep features, we use a continuous volumetric field of density and radiance. Different from the original NeRF\cite{mildenhall2021nerf}, for every queried 3D position \( x \in \mathbb{R}^3 \), we get a volume density \(\sigma(x)\) and a multi-channel feature \( F(x) \in \mathbb{R}^C \) instead of an RGB color, where \(C\) is the number of the feature channels. Then we can get the feature of any rays \(\textbf{r}\) passing through the volume by integrating sampled points along the ray via approximated volume rendering \cite{mildenhall2021nerf}:
\begin{equation}
    F(\textbf{r}) = \sum_{i=1}^N w_i F_i,
    \label{eq:volumerender}
\end{equation}
\begin{equation}
    \text{where}\quad w_i = \mathrm{exp} \left(-\sum_{j=1}^{i-1} \sigma_j \delta_j    \right) \left( 1-\mathrm{exp}\left( -\sigma_i \delta_i \right) \right),
    \label{eq:weight}
\end{equation}
where 
\(\sigma_i, F_i\) denotes the volume density and feature of sampled point \(i\), \(w_i\) denotes the weight of \(F_i\) in the ray \(\textbf{r}\),and \(\delta_i\) is the distance between adjacent samples. We disable view-dependency effect for better multi-view consistency.

Then we can generate feature maps capturing high-level features and map them to RGB space using a 2D CNN decoder. However, unlike \cite{chan2022efficient,niemeyer2021giraffe,gu2021stylenerf}, we render full-resolution feature maps which have the same resolution as the final RGB images rather than down-sampled feature maps. Rendering full-resolution feature maps has two unique features: \textbf{1)} it discards up-sampling operations which cause multi-view inconsistency in general \cite{gu2021stylenerf}, \textbf{2)} it removes aliasing when rendering low-resolution feature maps \cite{barron2021mip} which causes severe flickering effects in stylized RGB videos.

Directly using 3D voxel grid to store features is memory-intensive. We thus adopt vector-matrix tensor decomposition\cite{chen2022tensorf} that relaxes the low-rank constraints for two modes of a 3D tensor and factorizes tensors into compact vector and matrix factors, which lowers the space complexity from \(\mathcal{O}(n^3)\) to \(\mathcal{O}(n^2)\), massively reducing the memory footprint. We employ a density grid to store volume density and a feature grid to store multi-channel features respectively.

%% file: sections/styletransfer.tex
\subsection{Feature Transformation for Style Transfer}
\label{sec:styletransfer}

Once we have the feature grid representation of a scene, we can tackle the task of stylizing 3D scenes. Given a reference style image, our goal is to render stylized novel views of the 3D scene with multi-view consistency. To achieve this, we apply transformations to the features of the grid. 

One plausible solution to this task is to apply style transfer to the feature grid directly. This solution is efficient in evaluations as it can render any stylized views with a single style transfer process only. However, it is impractical to train such transformation as it needs to stylize the whole feature grid in every iteration. Another solution is to apply an off-the-shelf zero-shot style transfer method to the features of the sampled 3D points. While this solution can reduce computational cost through decreasing the size of training patch and the number of sampled points, it has two problems:  \textbf{1)}  vanilla zero-shot style transformation is conditioned on holistic statistics of the sampled point batch \cite{li2019learning, huang2017arbitrary, liu2021adaattn}, which violates multi-view consistency in volume rendering as the feature transformation of a specific 3D point will vary across different sampled points; \textbf{2)} volume rendering requires sampling hundreds of points along a single ray, which makes transformation on the point batch memory-intensive.

Motivated by the observation that style transformation is conditioned on both content information and style information, we decompose the style transformation into sampling-invariant content transformation (SICT) and deferred style transformation (DST). After the decomposition, SICT will be conditioned solely on the content information while DST conditioned solely on the style information, more details to be elaborated in the ensuing subsections.

\vspace{-0.5em}
\subsubsection{Sampling-invariant Content Transformation}
\label{sec:SICT}

\begin{figure}
    \hfill
    \begin{subfigure}[b]{0.4\linewidth}
    \includegraphics[scale=.65]{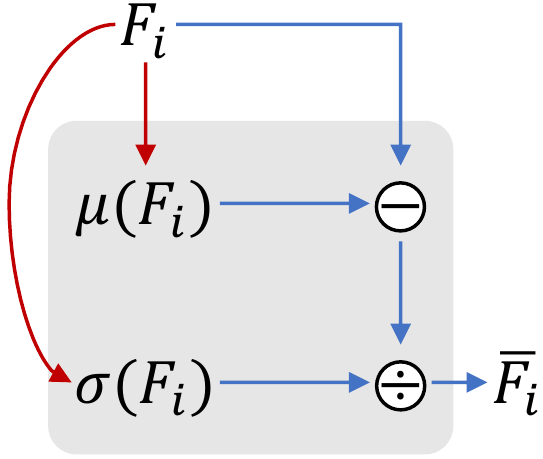}    \caption{Vanilla IN.}
    \end{subfigure}
    \hfill
    \begin{subfigure}[b]{0.4\linewidth}
    \includegraphics[scale=.65]{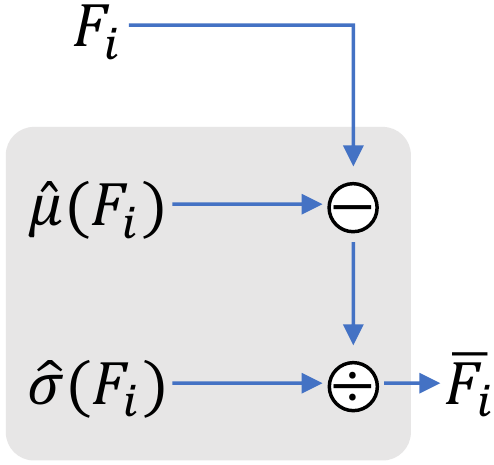}
    \caption{Volume-adaptive IN.}
    \end{subfigure}
    \hfill
    \caption{Comparison between vanilla instance normalization (IN) in (a) and volume-adaptive 
    IN in (b). During evaluation, volume-adaptive 
    IN uses learned mean and standard-deviation, discarding dependency over the sampled point batch's holistic statistics (indicated by the \textcolor{red}{red} arrows in the left graph).}
    \vspace{-1.2em}
    \label{fig:IN}
\end{figure}

Given a batch of sampled points, we can get their corresponding features \(F_i \in \mathbb{R}^{C}, i \in [1,2,...,N] \) from the feature grid, where \(N\) is the number of the sampled points along a ray and \(C\) is the number of the feature channels. The goal of SICT is to transform the extracted features \(F_i\) so that they can be better stylized. We formulate SICT as a channel-wise self-attention operation to the features after instance normalization (IN) \cite{ulyanov2016instance}. Specifically, we formulate \(Q\)(query), \(K\)(key), and \(V\)(value) as:
\begin{align}
    Q &= q(Norm(F_i)), \\
    K &= k(Norm(F_i)), \\
    V &= v(Norm(F_i)),
\end{align}
where \(q,k,v\) are \(1\times 1\) convolution layers which reduce the channel number from \(C\) to \(C'\) for computational efficiency, and \(Norm\) denotes the 
IN. However, as shown in \cref{fig:IN}, vanilla 
IN calculates per-dimension mean and standard-deviation of the batch of sampled points, which varies with different sampled points and incurs multi-view inconsistency accordingly. Thus we design volume-adaptive 
IN which, during training, keeps running estimates of the computed mean and standard-deviation, and uses them for normalization during evaluations (instead of computing from the sampled point batch). Through volume-adaptive 
IN, we can ensure that the content transformation is consistent regardless of the sampled point batch's holistic statistics.

Channel-wise self-attention can thus be implemented by:

\begin{equation}
    \bar{F_i} = V \otimes \mathrm{Softmax}\left( \widetilde{cov}(Q,K)\right),
\end{equation}
where  \(\otimes\) denotes matrix multiplication and \(\widetilde{cov}(Q,K) \in \mathbb{R}^{N\times C' \times C'}\) denotes the covariance matrix in the channel dimension.

\vspace{-0.5em}
\subsubsection{Deferred Style Transformation}
\label{sec:DST}

\begin{figure}
    \centering
    \includegraphics[scale=0.9]{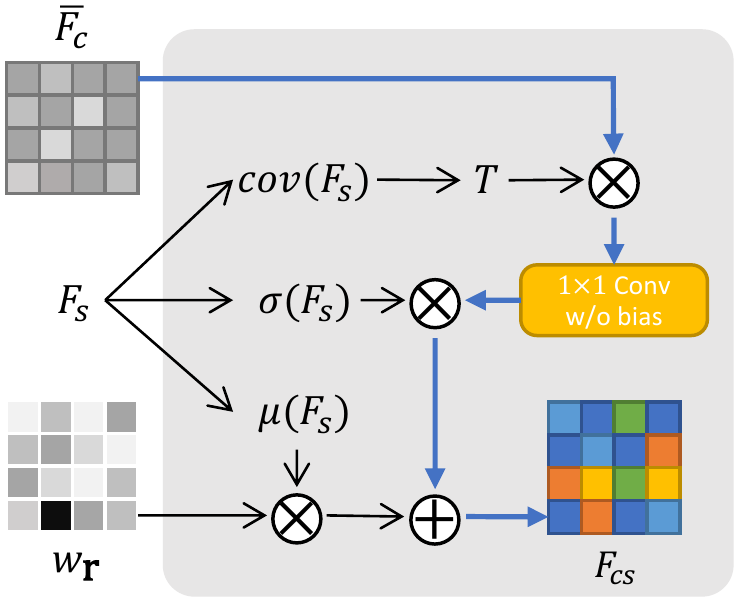}
    \caption{\textbf{Deferred style transformation.}  We apply the style transformation to the volume-rendered feature maps \(\bar{F_c}\) according to the style feature maps \(F_s\). To ensure multi-view  consistency, we modulate the bias (e.g. the mean value of the style feature maps \(\mu(F_s)\))  with the sum weight of sampled points along each ray \(w_{\textbf{r}}\).}
    \vspace{-1.2em}
    \label{fig:DST}
\end{figure}

After applying SICT to the features of each 3D point, we apply DST to the volume-rendered 2D feature maps \(\bar{F_c}\) rather than 3D point features \(\bar{F_i}\). To ensure multi-view consistency, we formulate the transformation as matrix multiplication and adaptive bias addition as illustrated in \cref{fig:DST}.

Specifically, we first extract feature maps \(F_s\) of the reference style \(S\) using a pre-trained VGG\cite{simonyan2014very}, and then generate the style transformation matrix \(T \in \mathbb{R}^{C'\times C'}\) using feature covariance \(cov(F_s)\) following \cite{li2019learning}. Next, we apply matrix multiplication with \(T\) to the feature maps \(\bar{F_c}\) and use a \(1 \times 1\) convolution layer \(conv\) without bias to restore the channel number from \(C'\) to \(C\). Though these operations can partially instill style information, they are not expressive enough without bias addition containing style information \cite{wu2021styleformer}. Thus following \cite{huang2017arbitrary}, we multiply the feature maps with the standard-deviation value \(\sigma(F_s) \) and add the mean value \(\mu(F_s)\). To ensure it is equivalent when applying the transformation to either 3D point features or 2D feature maps, we adaptively modulate the mean value \(\mu(F_s)\) with the sum weight of sampled points along each ray \(w_{\textbf{r}}\). DST can be mathematically formulated by:
\begin{equation}
    F_{cs} = conv \left(  T \otimes \bar{F_c}  \right)  \times \sigma(F_s) + w_{\textbf{r}} \times \mu(F_s),
    \label{eq:styletrans}
\end{equation}
\begin{equation}
    \text{where}\quad \bar{F_c} = \sum_{i=1}^N w_i \bar{F_i}, w_{\textbf{r}} = \sum_{i=1}^N w_i, \textbf{r} \in \mathcal{R}
\end{equation}
where \(w_i\) denotes the weight of sampled point \(i\) (\cref{eq:weight}), \(\bar{F_i}\) denotes the feature of sample \(i\) after SICT, and \(\mathcal{R}\) is the set of rays in each training batch.

Note \(conv\) is a \(1 \times 1\) convolution layer without bias, so it is basically a matrix multiplication operation. And \(\sigma(S), \mu(S)\) are scalars. Together with the adaptive bias modulation \(w_{\textbf{r}}\), \cref{eq:styletrans} can be reformulated by:
\begin{equation}
    F_{cs} = \sum_{i=1}^N w_i \left(  \underbrace{  conv\left( T \otimes \bar{F_i} \right) \times \sigma(F_s)  + \mu(F_s)   }_{\mbox{(i)}} \right),
    \label{eq:styletrans2}
\end{equation}
where part \(\mbox{(i)}\) can be seen as applying style transformation on every 3D point feature independently before volume rendering. This proves that applying DST on 2D feature maps is equivalent to applying the transformation on 3D points’ features, maintaining multi-view consistency. The full derivation of \cref{eq:styletrans2} is provided in the appendix.

Finally, we adopt a 2D CNN decoder to project the stylized feature maps \(F_{cs}\) to RGB space to generate the final stylized novel view images.

%% file: sections/training.tex
\subsection{Two-stage Model Training}
The training of our model is divided into the \emph{feature grid training stage} and the \emph{stylization training stage}, the former is trained with the target of novel view synthesis, and the latter is trained with the target of style transfer.

\noindent
\textbf{Feature grid training stage (First stage). }
We first learn the feature grid 3D representation for the novel view synthesis task, in preparation for performing feature transformation for style transfer. We train the feature grid and the 2D CNN decoder simultaneously, with the supervision of both RGB images and their bilinearly up-sampled feature maps extracted from \verb+ReLU3_1+ layer of pre-trained VGG\cite{simonyan2014very}. By aligning the VGG features with the feature grid, the reconstructed features acquire semantic information.
We use density grid pre-trained solely on RGB images since the supervising feature maps are not strictly multi-view consistent. The training objective is the mean square error (MSE) between the predicted and ground truth feature maps and RGB images. Following \cite{mu20223d,huang2021learning}, we use perceptual loss\cite{johnson2016perceptual} as additional supervision to increase reconstructed image quality. The overall loss function is:
\begin{multline}
    \mathcal{L}_{grid} =  \sum_{r \in \mathcal{R}}  \left \| \hat{F}(\textbf{r}) - F(\textbf{r})  \right \|_2^2 + \left \| \hat{I}_{\mathcal{R}} - I_{\mathcal{R}}  \right \|_2^2 \\
    + \sum_{l\in l_p} \left \| {\cal{F}}^l( \hat{I}_{\mathcal{R}} ) - {\cal{F}}^l(I_{\mathcal{R}}) \right \|_2^2 ,
\end{multline}
where \(\mathcal{R}\) is the set of rays in each training batch, \( \hat{F}(\textbf{r}), F(\textbf{r})\) are the predicted and ground truth feature of ray \(\textbf{r}\), \(\hat{I}_{\mathcal{R}}, I_{\mathcal{R}}\) are the predicted and ground truth RGB image,  \(l_p\) denotes the set of VGG layers 
that compute perceptual loss, ${\cal{F}}^l$ denotes the feature maps 
of the \(l\)th layer of pre-trained VGG network.

\noindent
\textbf{Stylization training stage (Second stage). } 
Our model learns to stylize novel views in the second stage. We freeze the feature grid, train the style transfer module, and fine-tune the CNN decoder.
Thanks to the memory-efficient representation of 3D scenes and DST, unlike \cite{schwarz2020graf, chiang2022stylizing, fan2022unified}, our model can be trained directly on \(256 \times 256\) patches, making patch sub-sampling algorithm \cite{huang2022stylizednerf, chiang2022stylizing, fan2022unified,schwarz2020graf} unnecessary. We use the same loss as \cite{huang2017arbitrary} where the content loss $\mathcal{L}_{c}$ is the MSE of the feature maps and the style loss $\mathcal{L}_{s}$ is the MSE of the channel-wise feature mean and standard-deviation: 
\begin{equation}
    \mathcal{L}_{stylization} = \mathcal{L}_{c} + \lambda\mathcal{L}_{s} ,
\end{equation}
where \(\lambda\) balances the content preservation and the stylization effect.

%% file: sections/results.tex
\begin{figure*}
    \centering
    \includegraphics[scale=.75]{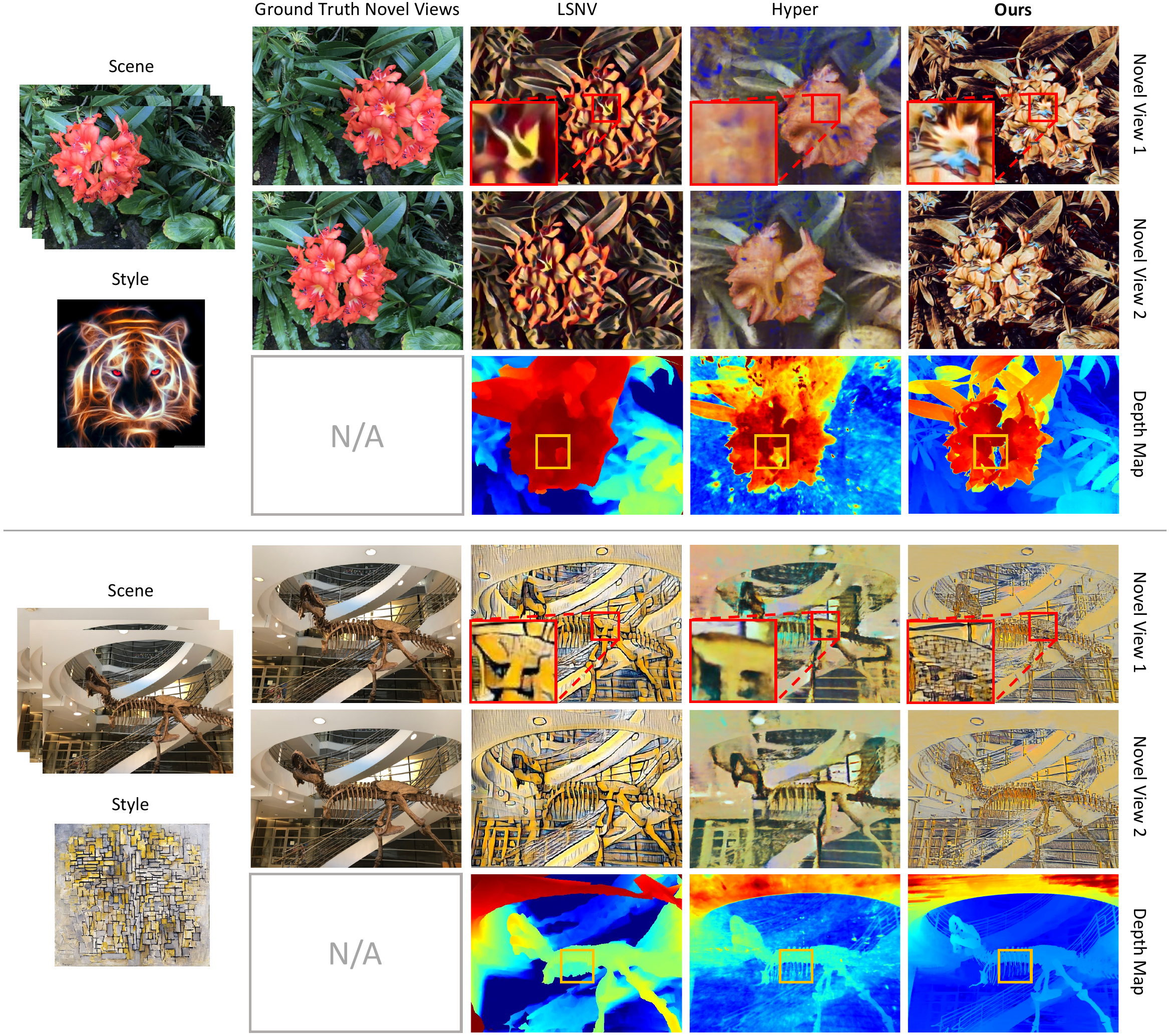}
    \caption{Comparison of StyleRF with two state-of-the-art zero-shot 3D style transfer methods LSNV \cite{huang2021learning} and Hyper \cite{chiang2022stylizing}. For each of the two sample \textit{Scenes} and reference \textit{Styles}, StyleRF produces clearly better 3D style transfer and depth estimation. Check zoom-in for details.}
    \vspace{-1.2em}
    \label{fig:comparison}
\end{figure*}

\section{Experiments}

We evaluate StyleRF extensively with qualitative experiments in \cref{sec:qualitative}, quantitative experiments in \cref{sec:quantitative} and ablation studies in  \cref{sec:ablation}. We demonstrate two applications of StyleRF in \cref{sec:app}. The implementation details are provided in the appendix.

\subsection{Qualitative Experiments}
\label{sec:qualitative}
We evaluate StyleRF over two public datasets including LLFF \cite{mildenhall2019llff} that contains real scenes with complex geometry structures and Synthetic NeRF \cite{mildenhall2021nerf} that contains \(360^{\circ}\) views of objects. In addition, we benchmark StyleRF with two state-of-the-art zero-shot 3D style transfer methods LSNV\cite{huang2021learning} and Hyper \cite{chiang2022stylizing} with their released codes. We perform comparisons on LLFF dataset \cite{mildenhall2019llff}.

\cref{fig:comparison} shows qualitative comparisons. We can see that StyleRF achieves clearly better stylization with more precise geometry reconstruction. Specifically, StyleRF can generate high-definition stylization with realistic textures and patterns of style images. The superior stylization is largely attributed to our transformation design that allows working in the feature space with full-resolution feature maps. As illustrated in the highlight boxes, StyleRF can successfully restore the intricate geometry of complex scenes thanks to its radiance field representations. In addition, only StyleRF faithfully transfers the squareness texture in the second style image. Furthermore, StyleRF can robustly generalize to new styles in a zero-shot manner and can adapt well to \(360^{\circ}\) dataset as illustrated in \cref{fig:teaser}. As a comparison, LSNV \cite{huang2021learning} fails to capture fine-level geometry like the bones of the T-Rex and the petals of the flower while Hyper \cite{chiang2022stylizing} produces very blurry stylization.

\subsection{Quantitative Results}
\label{sec:quantitative}

\begin{table}


\centering

\resizebox{\linewidth}{!}{
    \begin{tabular}{*5c}
    \toprule
    Method & \multicolumn{2}{c}{\begin{tabular}{@{}c@{}}Short-range \\ Consistency\end{tabular}} & \multicolumn{2}{c}{\begin{tabular}{@{}c@{}}Long-range \\ Consistency\end{tabular}} \\
    \midrule
    {}    & \textbf{LPIPS} & \textbf{RMSE}  & \textbf{LPIPS} & \textbf{RMSE} \\
    
    AdaIN\cite{huang2017arbitrary}  &  0.152  & 0.123  & 0.220  & 0.186   \\
    CCPL\cite{wu2022ccpl}           &  0.110  & 0.106  & 0.191  & 0.174   \\
    ReReVST\cite{ReReVST2020}       &  0.098  & 0.080  & 0.186  & 0.146   \\
    LSNV\cite{huang2021learning}    &  0.093  & 0.092  & 0.181  & 0.155   \\
    Hyper\cite{chiang2022stylizing} &  0.084  & 0.068  & 0.131  & 0.101    \\
    \textbf{Ours}                   &  0.072  & 0.082  & 0.149  & 0.137  \\
    \bottomrule
    \end{tabular}
}
\caption{\textbf{Results on consistency.}  We compare StyleRF with the state-of-the-art on consistency using LPIPS (\(\downarrow\)) and RMSE (\(\downarrow\)).}
\vspace{-1.2em}
\label{tab:compare}
\end{table}

3D style transfer is a very new and under-explored task and there are few metrics for quantitative evaluation of stylization quality. Hence, we manage to evaluate the multi-view consistency only. In our experiments, we warp one view to the other according to the optical flow \cite{teed2020raft} using softmax splatting \cite{niklaus2020softmax}, and then computed the masked RMSE score and LPIPS score \cite{zhang2018unreasonable} to measure the stylization consistency. Following \cite{huang2021learning,fan2022unified,chiang2022stylizing}, we compute the short-range and long-range consistency scores which compare adjacent views and  far-away views respectively. 
We compare StyleRF against two state-of-the-art zero-shot 3D style transfer methods Hyper \cite{chiang2022stylizing} and LSNV \cite{huang2021learning},  one SOTA single-frame-based video style transfer method CCPL\cite{wu2022ccpl}, one SOTA multi-frames-based video style transfer method ReReVST \cite{ReReVST2020}, and one classical image style transfer method AdaIN\cite{huang2017arbitrary}.

It can be seen from \cref{tab:compare} that StyleRF significantly outperforms image style transfer approach~\cite{huang2017arbitrary} and video style transfer approach~\cite{wu2022ccpl, ReReVST2020} which capture little information about the underlying 3D geometry. In addition, StyleRF achieves better consistency than point-cloud-based 3D style transfer \cite{huang2021learning} as well. Note Hyper\cite{chiang2022stylizing} achieves slightly better LPIPS and RMSE scores than our method, largely because it produces over-smooth results and inadequate stylization as shown in \cref{fig:comparison}.

%% file: sections/studies.tex
\subsection{Ablation Studies}
\label{sec:ablation}

\begin{figure}
    \centering
    \includegraphics[scale=.3]{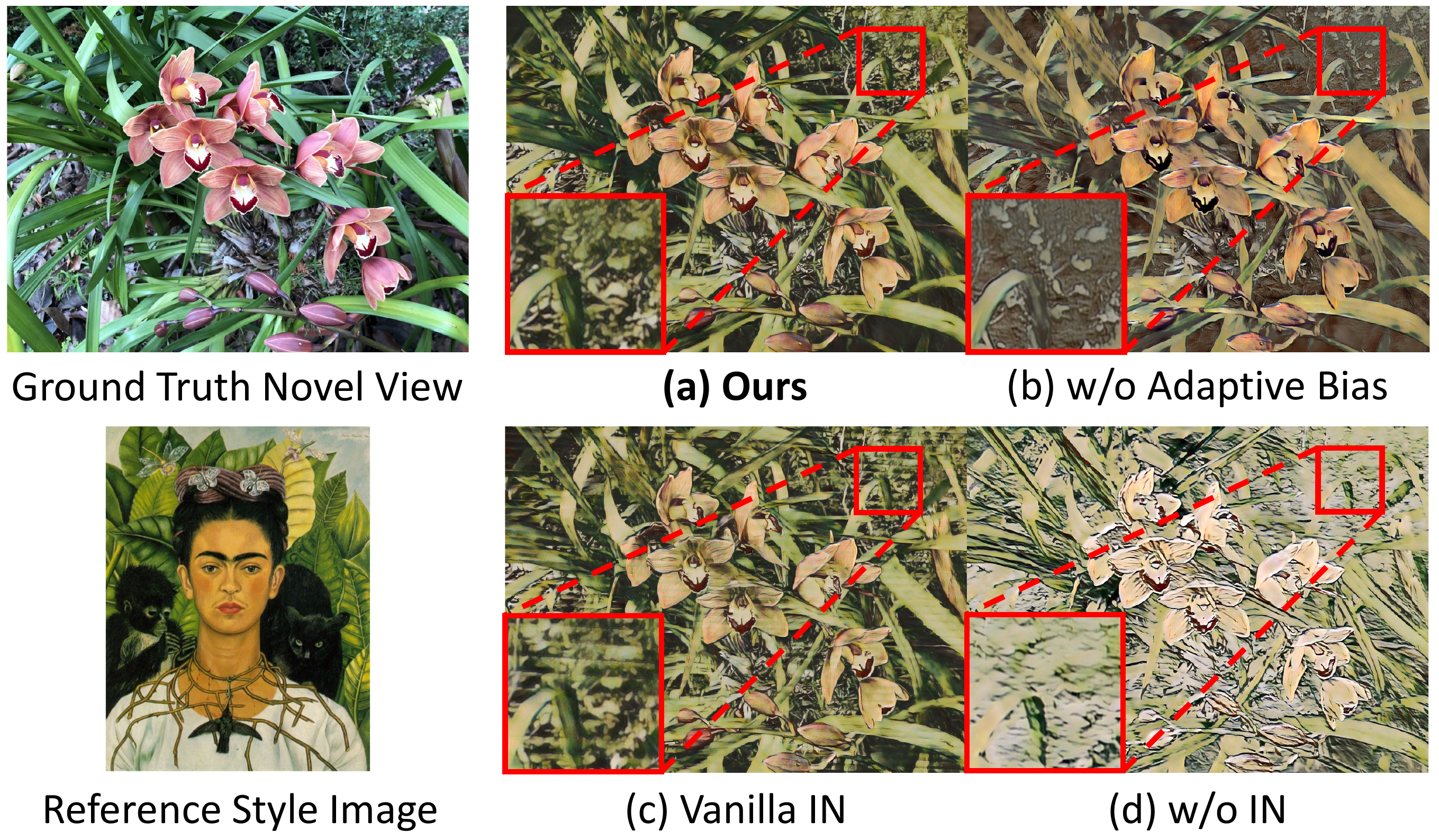}
    \caption{\textbf{Ablation studies.}  (a) shows the stylization of our full pipeline. (b) shows the stylization without the adaptive bias. (c) shows the stylization when replacing the volume-adaptive instance normalization (IN) with vanilla IN. (d) shows the stylization without any IN.}
    \vspace{-1em}
    \label{fig:ablation}
\end{figure}

We design two innovative techniques to improve the stylization quality and maintain the multi-view consistency. The first is volume-adaptive instance normalization which uses the learned mean and variance of the whole volume during inference, eliminating the dependency on holistic statistics of the sampled point batch. The second is the adaptive bias addition in DST, which improves the stylization quality using bias capturing style information. We evaluate the two designs to examine how they contribute to the overall stylization of our method.

\noindent \textbf{Volume-adaptive instance normalization.} 
We compare our volume-adaptive instance normalization (IN) with vanilla IN and StyleRF without IN. As \cref{fig:ablation} (c) shows, vanilla IN produces severe block-shape artifacts as the transformation of each batch is conditioned on the holistic statistics of itself, thus each batch (i.e. block in the image) produces inconsistent stylization which leads to the artifacts. However, if we discard IN as shown in \cref{fig:ablation} (d), the multi-view consistency can maintain but the stylization quality compromises a lot, failing to capture the correct color tone of the reference style image. This is because IN removes the original style information of the content image which facilitates the transfer of the reference style \cite{huang2017arbitrary}.

\noindent \textbf{Adaptive bias addition.}
As illustrated in \cref{fig:ablation} (b), the stylization quality degrades a lot if we eliminate the adaptive bias addition in DST (\cref{sec:DST}), producing unnatural stylization compared to the stylization of our full pipeline in \cref{fig:ablation} (a). This is because bias usually contains crucial style information such as the overall color tone \cite{wu2021styleformer}. StyleRF employs bias addition that is adaptively modulated by the weight of each ray, improving the stylization quality and keeping multi-view consistency concurrently.

\subsection{Applications}
\label{sec:app}

\begin{figure}
    \centering
    \includegraphics[scale=.35]{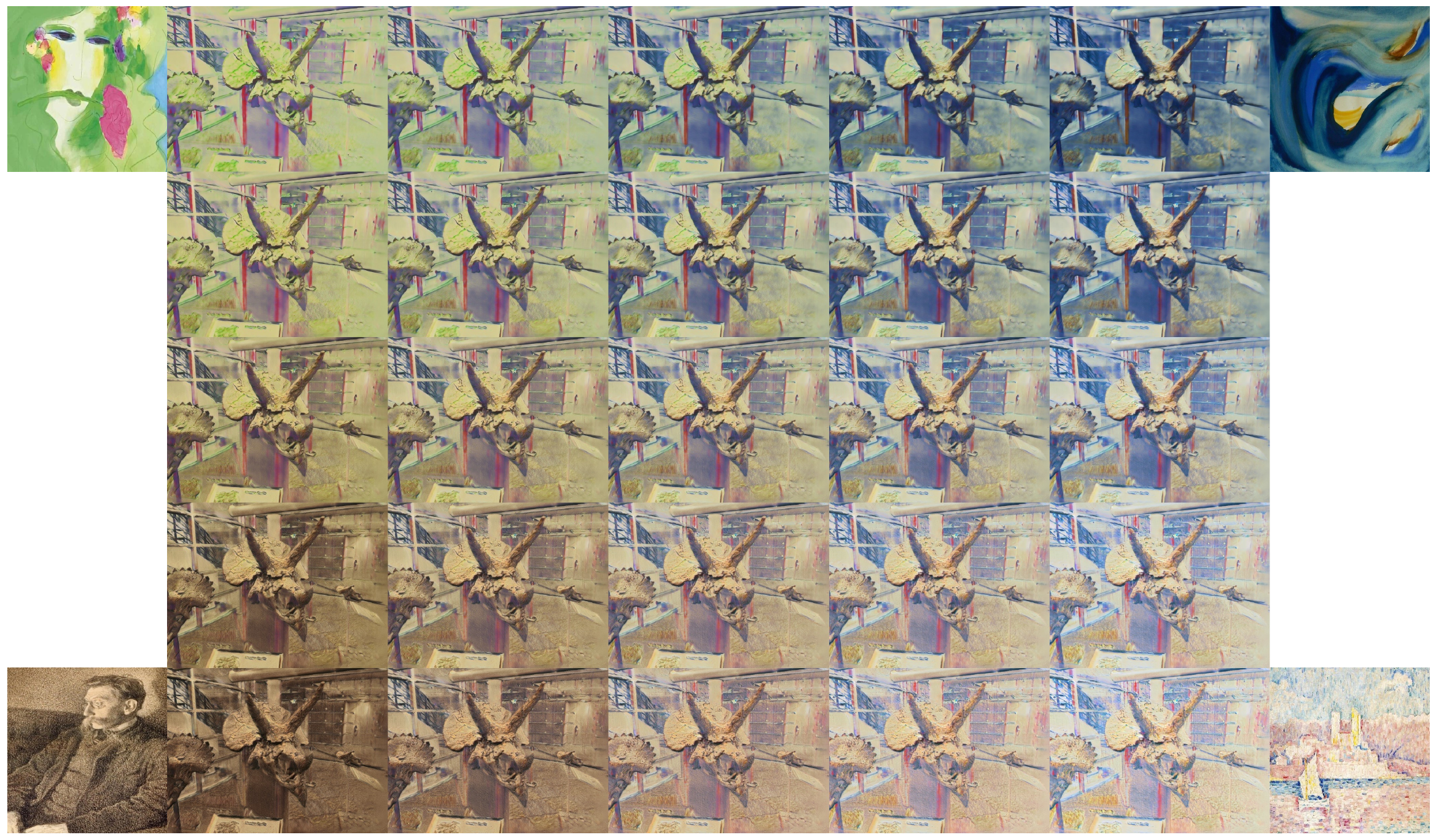}
    \caption{\textbf{Multi-style interpolation.}  StyleRF can smoothly interpolate between arbitrary styles by interpolating features of the scene.}
    \vspace{-1.2em}
    \label{fig:interpolation}
\end{figure}

\begin{figure}
    \centering
    \includegraphics[scale=.36]{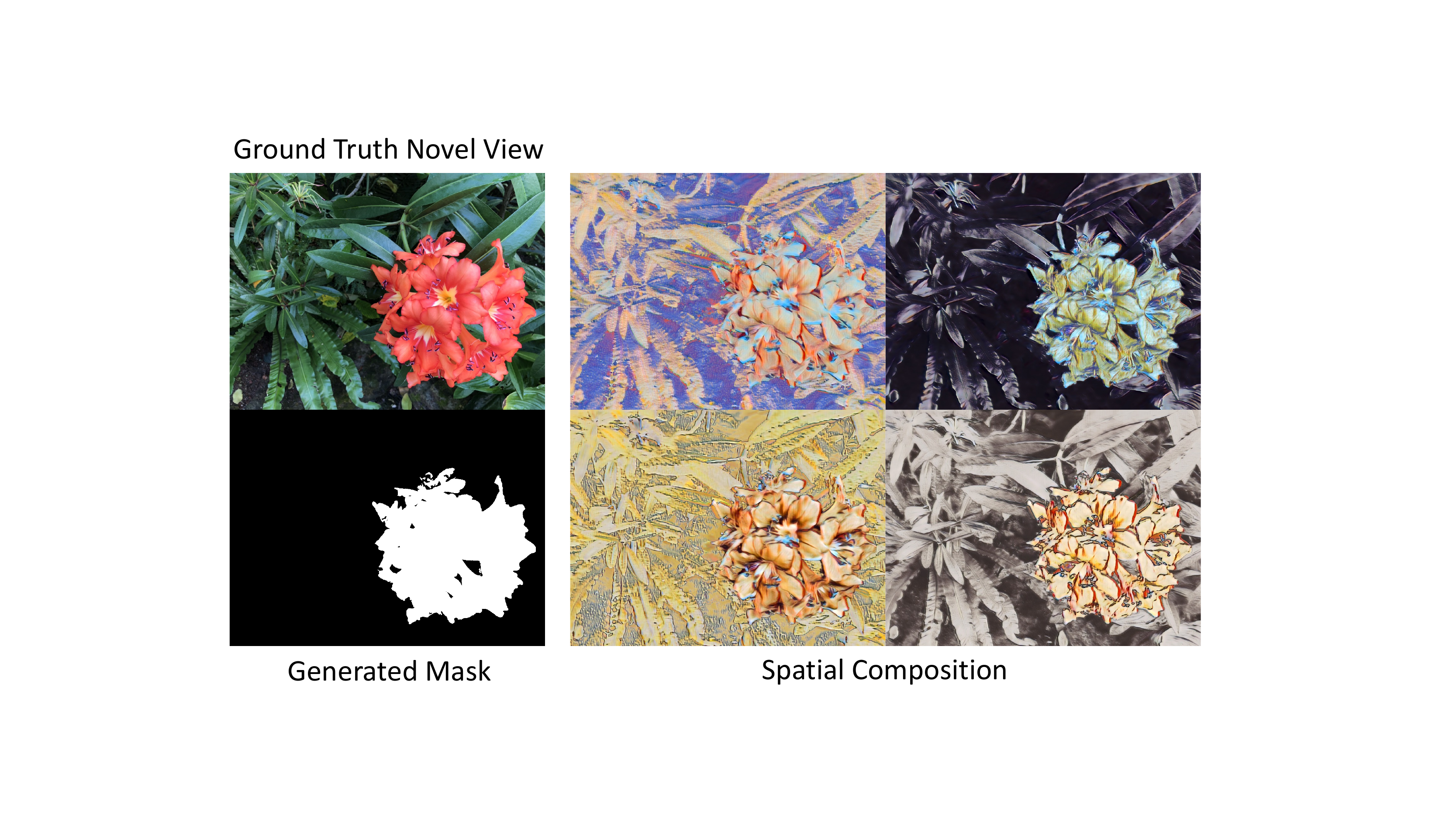}
    \caption{\textbf{Compositional 3D style transfer.}  Given the 3D-consistent segmentation masks, StyleRF can create infinite combinations of styles by spatial composition.}
    \vspace{-1.2em}
    \label{fig:composition}
\end{figure}

StyleRF can be easily extended along different directions with different applications. We provide two possible extensions in the ensuing subsections.    

\noindent \textbf{Multi-style interpolation.}
StyleRF can smoothly interpolate different styles thanks to its high-level feature representation of a 3D scene. As illustrated in \cref{fig:interpolation}, we linearly interpolate the feature maps of a specific view by using four different styles at four corners. Unlike previous NeRF-based 3D style transfer that supports style interpolation by interpolating one-hot latent vectors \cite{fan2022unified}, StyleRF can interpolate arbitrary numbers of unseen new styles by interpolating features of the scene, yielding more smooth and harmonious interpolation. Hence, StyleRF can not only transfer arbitrary styles in a zero-shot manner but also generate non-existent stylization via multi-style interpolation.

\noindent \textbf{Compositional 3D style transfer.}
Thanks to its precise geometry reconstruction, StyleRF can be seamlessly integrated with NeRF-based object segmentation \cite{fan2022nerf,Zhi:etal:ICCV2021,kobayashi2022decomposing} for compositional 3D style transfer. As shown in \cref{fig:composition}, we apply 3D-consistent segmentation masks to the feature maps and apply different styles to stylize the contents inside and outside the masks separately. We can see that the edges of the masks can be blended more softly by applying the segmentation masks to the feature maps instead of RGB images. Due to its zero-shot nature, StyleRF can create infinite combinations of styles without additional training, producing numerous artistic creations and inspirations.

%% file: sections/conclusion.tex
\section{Conclusion}
In this paper, we present StyleRF, a novel zero-shot 3D style transfer method that balances the three-way dilemma over accurate geometry reconstruction, high-quality stylization, and being generalizable to arbitrary new styles. By representing the 3D scene with an explicit grid of high-level features, we can faithfully restore high-fidelity geometry through volume rendering. Then we perform style transfer on the feature space of the scene, leading to high-quality zero-shot stylization results. We innovatively design sampling-invariant content transformation to maintain multiview consistency and deferred style transformation to increase efficiency. We demonstrate that StyleRF achieves superior 3D stylization quality than previous zero-shot 3D style transfer methods, and can be extended to various interesting applications for artistic 3D creations.

\subsection*{Acknowledgement}
\noindent
This project is funded by the Ministry of Education Singapore, under the Tier-1 project scheme with project number RT18/22.